\documentclass[letterpaper, 10 pt, conference]{ieeeconf} 
\IEEEoverridecommandlockouts                              
\usepackage{url}
\usepackage{cite}
\usepackage{float}
\usepackage{amsmath,amssymb,amsfonts}
\usepackage{hyperref}
\usepackage{cleveref}
\usepackage{graphicx}
\usepackage{textcomp}
\usepackage{booktabs}
\usepackage{multicol}
\usepackage{multirow}
\usepackage{subcaption}
\usepackage{xcolor}
\usepackage{cite}
\usepackage{tikz}
\usepackage{wrapfig}
\usepackage{xcolor}
\usepackage{layout}

\newtheorem{theorem}{Theorem}
\newtheorem{definition}{Definition}
\newenvironment{assumption}[1][Assumption]{\par\noindent\textbf{Assumption.}\itshape\ }{\par}
\crefname{assumption}{assumption}{assumptions}
\Crefname{assumption}{Assumption}{Assumptions}
\newenvironment{remark}{
  \par\noindent\textbf{Remark.}\itshape\ }{\par}

\def\BibTeX{{\rm B\kern-.05em{\sc i\kern-.025em b}\kern-.08em
    T\kern-.1667em\lower.7ex\hbox{E}\kern-.125emX}}

\definecolor{a_HA}{RGB}{255, 100, 100}  
\definecolor{a_HP}{RGB}{56, 100, 178}  

\definecolor{carnationpink}{rgb}{1.0, 0.65, 0.79}

\title{\LARGE \bf
Runtime Learning of Quadruped Robots in Wild Environments
}
\author{Yihao Cai$^{1}$, Yanbing Mao$^{1}$, Lui Sha$^{2}$,  Hongpeng Cao$^{3}$, Marco Caccamo$^{3}$, and Honghao Wei$^{4}$ 
\thanks{$^{1}$Yihao Cai and Yanbing Mao are with Engineering Technology Division, Wayne State University, Detroit, MI 48201, USA
{\tt\small \{yihao.cai, hm9062\}@wayne.edu}}%
\thanks{$^{2}$Lui Sha is with Department of
Computer Science, University of Illinois Urbana-Champaign, Champaign, IL, 61820, USA {\tt\small lrs@illinois.edu}}
\thanks{$^{3}$Hongpeng Cao and Marco Caccamo are with School of Engineering and Design, Technical University of Munich, Munich, 85748, Germany
{\tt\small \{cao.hongpeng, mcaccamo\}@tum.de}}%
\thanks{$^{4}$Honghao Wei is with School of Electrical Engineering and Computer Science, Washington State University, Pullman, WA 99163, USA
{\tt\small honghao.wei@wsu.edu}}%
}

\begin{document}

\maketitle

This paper presents a runtime learning framework for quadruped robots, enabling them to learn and adapt safely in dynamic wild environments. The framework integrates sensing, navigation, and control, forming a closed-loop system for the robot. The core novelty of this framework lies in two interactive and complementary components within the control module: the high-performance (HP)-Student and the high-assurance (HA)-Teacher. HP-Student is a deep reinforcement learning (DRL) agent that engages in self-learning and teaching-to-learn to develop a safe and high-performance action policy. HA-Teacher is a simplified yet verifiable physics-model-based controller, with the role of teaching HP-Student about safety while providing a backup for the robot's safe locomotion. HA-Teacher is innovative due to its real-time physics model, real-time action policy, and real-time control goals, all tailored to respond effectively to real-time wild environments, ensuring safety. The framework also includes a coordinator who effectively manages the interaction between HP-Student and HA-Teacher. Experiments involving a Unitree Go2 robot in Nvidia Isaac Gym and comparisons with state-of-the-art safe DRLs demonstrate the effectiveness of the proposed runtime learning framework. The corresponding open source code is available at \href{https://github.com/Charlescai123/isaac-runtime-go2}{\textcolor{blue}{github.com/Charlescai123/isaac-runtime-go2}}.

\section{Introduction}
Quadruped robots have become a promising solution for navigating challenging wild environments, such as forests, disaster zones, and mountainous regions  \cite{Lee_2020_terrain, agarwal2022leggedlocomotionchallengingterrains}. These robots are specifically designed to traverse unstructured terrains, slopes, and dynamic obstacles while maintaining stability and operational efficiency \cite{RobustRough-Terrain}. A typical example is their use in search and rescue tasks, where they assist in locating and aiding victims during earthquakes, building collapses, and other hazardous situations \cite{Mixed-reality}.

\subsection{Challenge and Open Problem} 
Deep reinforcement learning (DRL) has demonstrated considerable success in developing action policies that facilitate agile and efficient locomotion in quadruped robots \cite{tan2018simtoreallearningagilelocomotion, margolisyang2022rapid, yang2020data}. 
However, quadruped robots are typical safety-critical autonomous systems. A fundamental safety challenge emerges when implementing DRL-enabled locomotion strategies for quadruped robots in wild environments, detailed below. 

\noindent \textbf{Challenge: Dynamic Wild Environments.} The quadruped robots have dynamics-environments interactions, meaning their behavior and performance depend significantly on the terrains they traverse, such as flat ground versus uneven surfaces, and sandy terrain versus icy conditions. Furthermore, real-time wild environments are often unpredictable and can change unexpectedly, such as during freezing rain \cite{unf2}. These variations can create a domain gap for the trained locomotion policies, making it difficult for quadruped robots to operate safely and effectively in such environments. Therefore, it is essential to ensure that quadruped robots are resilient to these domain gaps, particularly those arising from dynamic and unpredictable wild environments.

An appealing prospect for addressing the aforementioned safety challenge is the DRL agent's runtime learning for adaptive action policies in wild environments. However, the open problems are \textit{If the DRL agent's actions lead to a safety violation, how can we correct its unsafe learning and guarantee robot's safety in a timely manner? How to adapt to dynamic wild environments for assuring safety?} Before presenting our approach to addressing them, we first review the existing related work.

\newlength{\customwidth}      
\newlength{\fighspace}        
\setlength{\customwidth}{0.14\textwidth}
\setlength{\fighspace}{-.245cm}
\begin{figure*}[http]
\vspace{-2.2cm}
\centerline{\includegraphics[width=1.025\textwidth]{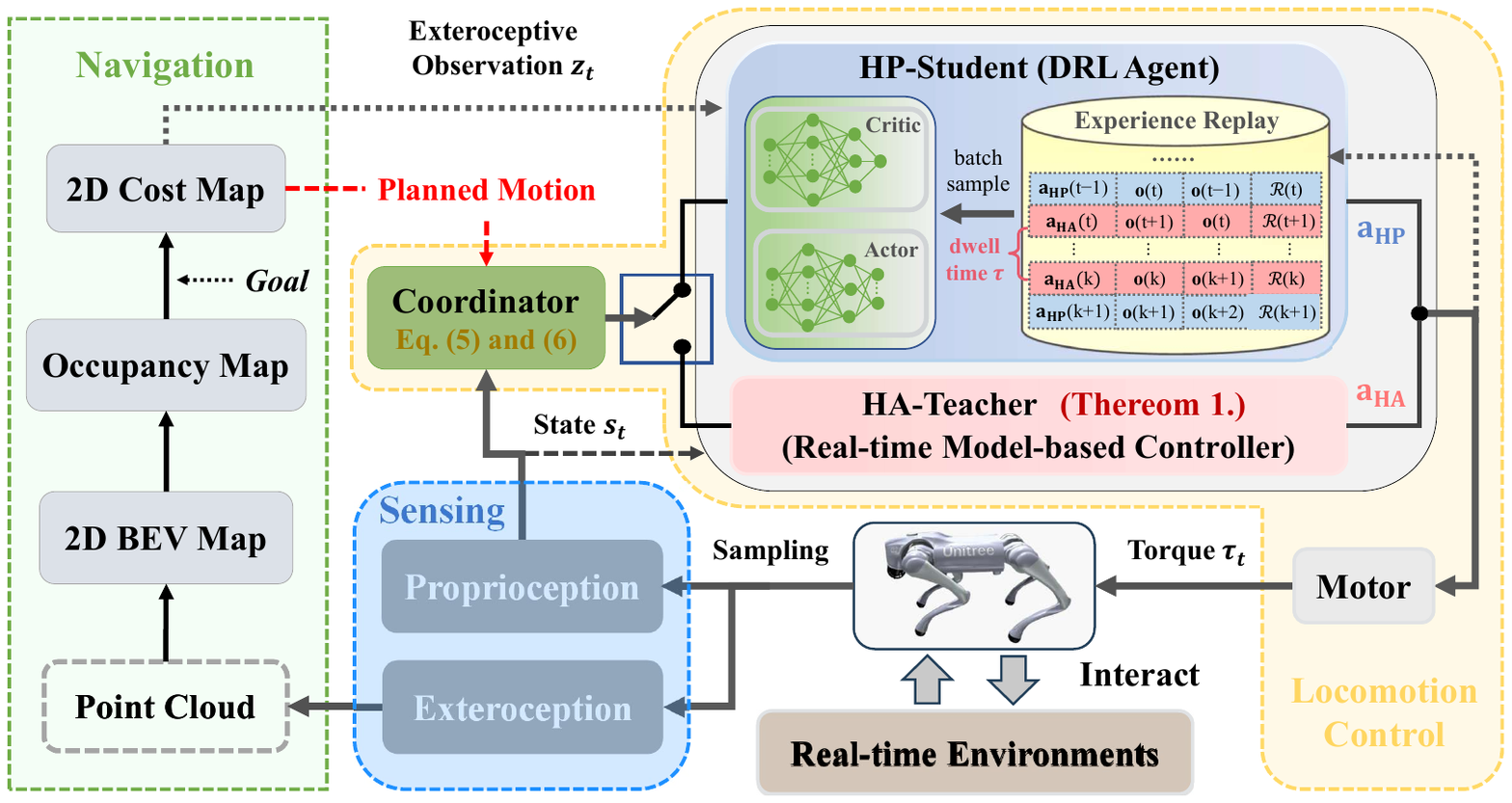}}
\vspace{-2.1cm}
\caption{Runtime Learning Framework: A seamless integration of perception, planning, learning, and control.} 
\vspace{0.1cm}
The \textbf{sensing} module outputs both proprioceptive and exteroceptive data from the robot operating in wild environments. The \textbf{navigation} module constructs a Bird's-Eye View (BEV) map by filtering Regions of Interest, generates an occupancy map using height thresholding, and computes the cost map with the Fast Marching method. Based on the cost map, a planner generates reference commands for the locomotion controller while also providing visual observations for the HP-Student. In the \textbf{locomotion control} module, the coordinator manages the interaction between the HA-Teacher and the HP-Student to ensure the robot's safety. This setup allows the HP-Student to learn from the HA-Teacher through $\textcolor{a_HA}{ \mathbf{a}_{\mathbf{HA}}}$ while also engaging in self-learning by $\textcolor{a_HP}{\mathbf{a}_{\mathbf{HP}}}$ to develop a safe and high-performance action policy. \label{learningframework}
\end{figure*}


\subsection{Related Work}
\textbf{DRL-enabled Locomotion}. Current DRL frameworks for quadruped robots typically involve pre-training locomotion policies in a source domain, such as a simulator, and then transferring these policies to a target domain, like the real world. The techniques used for addressing domain gaps during this transfer include zero-shot deployment \cite{kumar2021rma, tan2018simtoreallearningagilelocomotion}, fine-tuning \cite{funetune2022}, and domain randomization \cite{sadeghi2017cad2rl}, etc. However, these methods do not enable DRL's safe runtime learning in dynamic environments to have continuously adaptive DRL models after deployment. Consequently, they often struggle to handle unexpected environmental variations post-transfer. This limitation poses safety challenges, particularly in unpredictable wild environments that differ significantly from the training conditions in the source domain.

\textbf{Fault-tolerant DRL}. Recent approaches include neural Simplex \cite{phan2020neural} and runtime assurance \cite{brat2023runtime,sifakis2023trustworthy,chen2022runtime}. They treat the DRL agent as a high-performance module (HPM) but a black box that runs in parallel with a verified high-assurance module (HAM). Normally, HPM controls the real plants. HAM takes over once safety violation occurs. These architectures can ensure the safe running of DRL in real plants under the assumption that the real-time wild environments do not cause HAM to fail, which is not practical for quadruped robots in dynamic and unpredictable wild environments. Specifically, HAM is the static model-based controller, and its action will be unreliable if the real-time wild environments create a significant model mismatch for HAM design.

\subsection{Contribution: Runtime Learning Framework}
To address the aforementioned safety challenge associated with the quadruped robots operating in dynamic wild environments, we introduce our runtime learning framework shown in \cref{learningframework}, which integrate sensing, perception with planning, learning, and control. This framework enables quadruped robots to autonomously navigate and safely learn in real-time wild environments. Beyond addressing these safety concerns, this framework also tackles fundamental open problems in DRL-enabled locomotion and fault-tolerant DRL. Its core novelties are outlined below.

\begin{itemize}
    \item HP-Student -- a DRL agent, which performs both self-learning and learning-from-HA-Teacher for robots operating in real-time wild environments, targeting a safe and high-performance action policy.
    \item HA-Teacher -- a simplified yet verifiable controller based on physics models, designed solely to ensure safety. Its mission is to teach HP-Student about safe policies while backing up the robot's safe control. HA-Teacher utilizes a \textit{real-time} physics model, \textit{real-time} action, and \textit{real-time} control goals, all tailored to respond effectively to complex and dynamic environments, ensuring robot's runtime safety.
\end{itemize}
The interaction between HP-Student and HA-Teacher operates as follows: When the real-time status of the robot controlled by HP-Student approaches the safety boundary, HA-Teacher intervenes to enforce the robot’s safe control. During this period, HP-Student will learn from the HA-Teacher regarding the safe action policies (i.e., teaching-to-learn). Once the robot’s real-time status recovers from the safety boundary, control is handed back to HP-Student for continuing its self-learning. 
This interactive design fosters HP-Student's learning of a safe and high-performance policy under dynamic wild environments.

\thispagestyle{empty}
\pagestyle{empty}
\section{Preliminaries} 
\subsection{Notation} 
We use $\mathbf{P} \succ 0$ to represent that $\mathbf{P}$ is a positive definite matrix. $\mathbb{R}^{n}$ indicates the set of $n$-dimensional real vectors, while $\mathbb{N}$ denotes the set of natural numbers. The superscript `$\top$' indicates matrix transposition. We define $\mathbf{0}_{m}$ as an $m$-dimensional zero vector and $\mathbf{I}$ as the identity matrix of appropriate dimensions. The notation $\mathbf{c} > \mathbf{0}_{n}$ implies that the vector $\mathbf{c} \in \mathbb{R}^{n}$ is positive, meaning all its elements are strictly positive. Lastly, $\textbf{diag}\{\mathbf{c}\}$ is diagonal matrix whose diagonal entries are given by the elements of $\mathbf{c}$.

\subsection{Definitions} 
For quadruped robots, we define the following safety and action sets, to which both system states and action commands must always be constrained.
\begin{align}
&{\mathbb{S}} \triangleq \left\{ {\left. \mathbf{e} \in {\mathbb{R}^n} \right|-\mathbf{c} \le {\mathbf{C}} \cdot \mathbf{e}  \le \mathbf{c}},~~\text{with}~\mathbf{c} > \mathbf{0}_p  \right\}, \label{aset2}\\
&{\mathbb{A}} \triangleq \left\{ {\left. {\mathbf{a} \in {\mathbb{R}^m}} ~\right|~-\mathbf{d} \le {\mathbf{D}} \cdot \mathbf{a} \le \mathbf{d},~~\text{with}~\mathbf{d} > \mathbf{0}_q}\right\}, \label{org}  
\end{align}
where ${\mathbf{e}} \triangleq \mathbf{s} - \mathbf{r}$ denotes the tracking error of proprioceptive sampling $\mathbf{s}$ with respect to the planned motion $\mathbf{r}$ from the navigation module shown in \cref{learningframework}. Meanwhile, $\mathbf{C}$ and $\mathbf{c}$ are provided in advance to describe $p \in \mathbb{N}$ safety conditions, while $\mathbf{D}$ and $\mathbf{d}$ are used to describe $q \in \mathbb{N}$ conditions on physically-feasible action space for safe locomotion control.  The inequalities outlined in \cref{aset2} are sufficiently general to cover various safety conditions, such as robot's velocity regulation, yaw control for avoiding collisions, and management of the center of gravity to prevent falling.

\begin{figure}
\vspace{0.2cm}
\begin{center}
\includegraphics[width=0.45\textwidth]{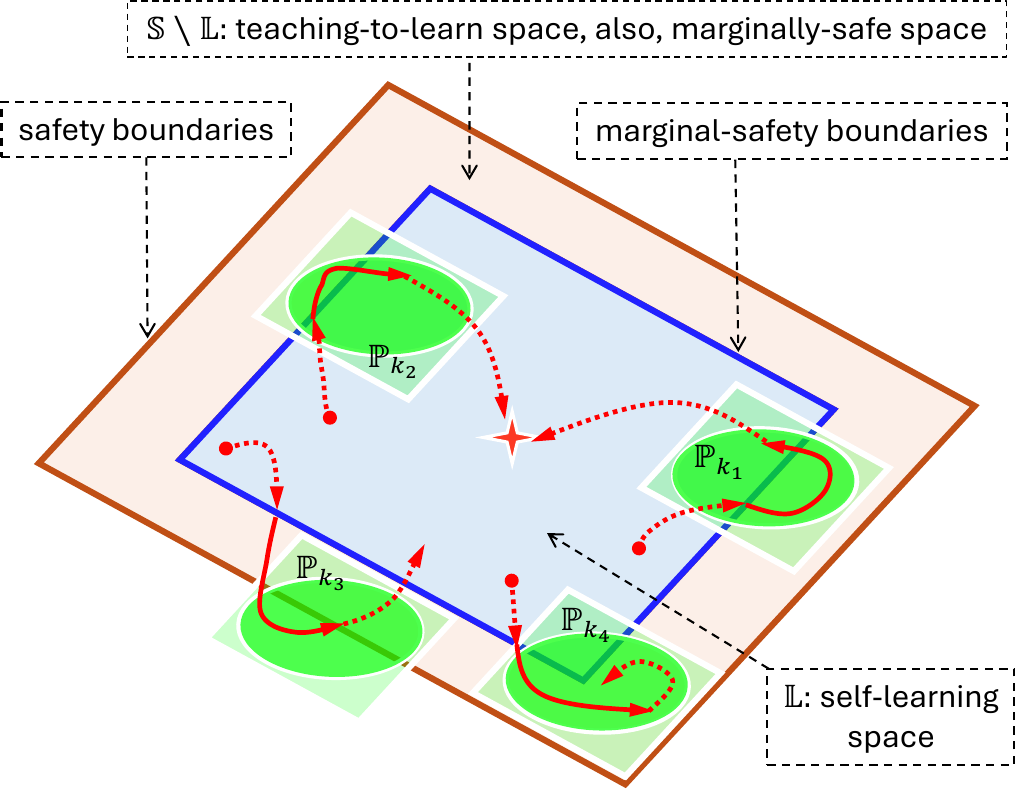}
\end{center}
\vspace{-0.3cm}
\caption{Illustrations of self-learning space, teaching-to-learn space (also referred to as marginally-safe space), real-time patches, safety boundaries, and marginal-safety boundaries. The $\mathbb{P}_{k_1}$ and $\mathbb{P}_{k_2}$ are successful real-time 
patches designed by \cref{thm10007p}, while $\mathbb{P}_{k_3}$ and $\mathbb{P}_{k_4}$ are failure cases since robot system under their control either leaves the safety set $\mathbb{S}$ (by $\mathbb{P}_{k_3}$) or cannot return to the self-learning space $\mathbb{L}$ (by $\mathbb{P}_{k_4}$).}
  \label{phasewithpatch}
  \vspace{-0.4cm}
\end{figure}

In our framework, the HP-Student (i.e., DRL agent) will engage in self-learning and teaching-to-learn for a safe and high-performance action policy. To understand them better, we refer to the safety set \eqref{aset2} and introduce the self-learning space of HP-Student:  
\begin{align} 
\mathbb{L} \triangleq \left\{ {\left. \mathbf{e} \in {\mathbb{R}^n} \right|-\eta \cdot \mathbf{c} \le {\mathbf{C}} \cdot \mathbf{e}  \le \eta \cdot \mathbf{c}},~0 < \eta < 1  \right\}. \label{aset3}
\end{align}
Given $0 < \eta < 1$, and $\mathbf{c} > \mathbf{0}_p$, it follows directly from \cref{aset2,aset3} that the self-learning space is a subset of the safety set, i.e., ${\mathbb{L}} \subset {\mathbb{S}}$. The intentional gap between these two sets accounts for the system’s response time and decision-making latency, which stem from physical constraints, computational limitations, and communication delays. 
This means when system states arrive at a safety boundary, action policy cannot immediately drive it back to the safety set, without time delay. 
As shown in \cref{phasewithpatch}, the boundaries of the self-learning space $\mathbb{L}$ can thus be regarded as the marginal-safety boundaries. Based on the set, we introduce the definition of HP-Student's safe action policy. 
\begin{definition}[Safe Action Policy of HP-Student]
Consider the self-learning space $\mathbb{L}$ \eqref{aset3}. The action policy of the DRL-agent (i.e., HP-Student) is said to be safe if, under its control, the robot's state satisfies that given $\mathbf{e}(1) \in \mathbb{L}$, the $\mathbf{e}(t) \in \mathbb{L}$ holds for all time steps $t \in \mathbb{N}$.
\label{hpsafe}
\end{definition}

In our framework \cref{learningframework}, the HA-Teacher serves as a physics-model-based controller dedicated solely to ensuring safety. When HP-Student's real-time actions are unsafe according to \cref{hpsafe}, the HA-Teacher will instruct the HP-Student about safety, fostering a ``teaching-to-learn" paradigm, while backup the robot's safe locomotion control simultaneously. As indicated in \cref{aset3,aset2,phasewithpatch}, the set $\mathbb{S}  \setminus  \mathbb{L}$ defines the HP-Student's teaching-to-learn space, also referred to as the marginally-safe space. We note the action policy of the HA-Teacher must have assured safety; otherwise, its safety teaching will mislead HP-Student. Therefore, we introduce the definition of ``assured safety," which guides the design of HA-Teacher later.
\begin{definition} [Safe Action Policy of HA-Teacher]
Consider safe set $\mathbb{S}$ \eqref{aset2}, action set $\mathbb{A}$ \eqref{org}, self-learning space $\mathbb{L}$ \eqref{aset3}, and set teaching horizon as $\tau \in \mathbb{N}$ and teaching period as 
\begin{align} 
\mathbb{T}^{\tau}_{k} \triangleq \{k+1, ~k+2, ~\ldots, ~k+\tau\}. \label{tp}
\end{align}
HA-Teacher's action policy denoted by $\pi_{\text{HA}}(\cdot)$ is said to have assured safety, if $\mathbf{e}(k) \in \mathbb{S}  \setminus  \mathbb{L}$, then i) 
the $\mathbf{e}(t) \in \mathbb{S}$ holds for any time $t \in \mathbb{T}^{\tau}_{k}$, ii) $\mathbf{a}_{\text{HA}}(t) = \pi_{\text{HA}}(\mathbf{e}(t)) \in \mathbb{A}$ holds for any time $t \in \mathbb{T}^{\tau}_{k}$, and iii)
$\mathbf{s}(k+\tau) \in \mathbb{L}$. \label{csafe}
\end{definition}

By \cref{csafe}, the HA-Teacher is deemed trustworthy for guiding the HP-Student in safety learning only if its action policy satisfies the following conditions:
i) it must ensure the robot consistently adheres to the safety regulations within $\mathbb{S}$; ii). it must keep the real-time actions within a physically-feasible action space $\mathbb{A}$; and iii) it must ensure the robot states return to the self-learning space $\mathbb{L}$ as teaching session ends.
Notably, if condition iii) is not met, the HA-Teacher will dominate, preventing the HP-Student from developing a high-performance action policy through self-learning.

\section{Runtime Learning Framework}
Referring to \cref{learningframework}, we describe the design of our runtime learning framework for enabling safe runtime learning in dynamic wild environments. 

\subsection{Sensing Module}
The sensing module provides both proprioceptive and exteroceptive data from the robot’s onboard sensors. Proprioceptive sensing delivers robot's state data that includes: 1) direct readings from the Inertial Measurement Unit, gyroscope, and motor encoders, which measure orientation and angular velocity; and 2) estimated center of mass (CoM) height (using a Kalman filter), and velocities in the CoM-x, CoM-y, and CoM-z directions. Exteroceptive sensing enables the perception of external terrain and obstacles. It creates 3D point cloud data using a depth camera, aiding in environmental mapping and obstacle avoidance for navigation.

\subsection{Navigation Module}
During runtime learning, the navigation module functions as a high-level motion planner, generating planned motions for the robot's locomotion controller to navigate wild environments. As shown in \cref{learningframework}, the pipeline includes: 1) constructing a Bird's-Eye View (BEV) map and an occupancy map; ii) applying the Fast Marching Method (FMM) to compute a 2D cost map; and iii) planning motions. 

\subsubsection{\textbf{2D Map Construction}}
The point cloud data is obtained from the quadruped robot's depth camera and transformed into the world frame. To enhance computational efficiency, we filter the point cloud using a Region of Interest, which allows us to retain only the relevant environmental features. The filtered data is then projected onto a discretized 2D occupancy grid, with each grid cell encoding local terrain characteristics. This process helps create the BEV map. Next, to further process the BEV representation, we generate a binary occupancy map based on a pre-defined maximum height of interest. In this map, any region that exceeds the height of the robot's body is classified as an obstacle. Within the occupancy map, we project the goal point from the world frame onto the map and apply the FMM to compute a 2D cost map, which will guide motion planning \cite{fu2022couplingvisionproprioceptionnavigation}. The 2D cost map generation pipeline can be found in \cref{ep5}.

\subsubsection{\textbf{Motion Planning}}
Based on the robot's position on the 2D cost map and its field of view, multiple short-term goals are identified. For each candidate goal, potential velocity sets are calculated, taking into account the robot's current velocity and control frequency. To ensure smooth movement, spline-based interpolation is employed to generate continuous motion references. The optimal reference is selected by minimizing the accumulated cost along the path on the map. Finally, this optimal motion reference is sent to the locomotion control module, allowing for seamless navigation with dynamically feasible motions.

\subsection{Locomotion Module}
This module consists of two components: HA-Teacher, which is a real-time physics-model-based controller, and HP-Student, which is a DRL agent. Their interactive design incorporates the core innovations of the proposed runtime learning framework. The motivation for developing this module stems from the two common approaches to achieving locomotion control in quadruped robots. The first approach is data-driven DRL, which delivers superior performance but poses challenges regarding verifiable safety. This is due to the vast number of parameters in DNNs, nonlinear activation functions, and various random factors. On the other hand, the physics-model-based controller (e.g., LQR) offers verifiable safety and stability, but its performance is often limited due to model mismatches. The characteristics of both approaches inspire us to integrate them, aiming to bring safe runtime learning into reality. Next, we detail the design. 

\subsubsection{\textbf{Coordinator}} As shown in \cref{learningframework}, the locomotion module has a coordinator, which is responsible for managing interactions between HP-Student and HA-Teacher through monitoring the condition:  
\begin{align}
\mathbf{e}(k-1) \in  \mathbb{L}~\text{and}~ \mathbf{e}(k) \in \mathbb{S}  \setminus  \mathbb{L}
\label{trigger}
\end{align}
by which, the data series in \cref{safedata} for teaching-to-learn in $\mathbb{S}  \setminus \mathbb{L}$ can be generated, and the switching logic of actions applied to the robot for safe locomotion is as follows:
\begin{align}
\mathbf{a}(t) \leftarrow \begin{cases}
		\mathbf{a}_{\text{HA}}(t), &\text{if condition}~\eqref{trigger}~\text{holds}~\text{and}~t \in \mathbb{T}_{k}\\ 
        \mathbf{a}_{\text{HP}}(t).      &\text{otherwise}  
	\end{cases}. 
\end{align}

\subsubsection{\textbf{HA-Teacher: Safety Only}} HA-Teacher's action policy is designed to be adaptive to real-time wild environments to assure safety only. Quadruped robots have dynamics-environment interactions. When a real-time environment creates safety issues, it is crucial to update the dynamics models, action policy, and control goals promptly to ensure safe and effective responses in real time. This insight has inspired us to develop a real-time patch for HA-Teacher: 
\begin{align}
\text{Patch:}~{\mathbb{P}_{k}} \triangleq &\left\{ {\left. \mathbf{e} \in {\mathbb{R}^n} \right|-\theta \cdot \mathbf{c} \le {\mathbf{C}} \cdot (\mathbf{e} - \mathbf{e}_{k}^*)  \le \theta \cdot \mathbf{c}},\right. \nonumber\\
&\hspace{3.9cm} \left. ~~0 < \theta < 1  \right\}, \label{aset4}
\end{align}
where $\mathbf{e}_{k}^*$ represents the patch center and serves as the real-time control goal, which is defined below.
\begin{align} 
\!\!\!\mathbf{e}_{k}^* \triangleq \chi \cdot \mathbf{e}(k), \!~\text{where}~0 \!<\! \chi \!<\! 1 ~\text{and}~\text{condition}~\eqref{trigger}~\text{holds}. \label{center}
\end{align} 
With the control goal at hand, we introduce the following real-time dynamics model of tracking errors w.r.t. $\mathbf{e}_{k}^*$, which is derived from the robot's dynamics model in \cite{di2018dynamic}.  
\begin{align}
&\widehat{\mathbf{e}}(t+1) = \mathbf{A}(\mathbf{s}(k)) \cdot \widehat{\mathbf{e}}(t) + {\mathbf{B}}(\mathbf{s}(k)) \cdot \mathbf{a}_{\text{HA}}(t) + \mathbf{h}(\widehat{\mathbf{e}}(t)), \nonumber\\
&\hspace{2.1cm}\text{for}~t \in \mathbb{T}^{\tau}_{k} \triangleq \{k+1, k+2, \ldots, k+\tau\} \label{realsyserror}
\end{align} 
where $\widehat{\mathbf{e}}(t) \triangleq \mathbf{e}(t) - \mathbf{e}_{k}^*$, $\mathbf{h}(\widehat{\mathbf{e}}(t))$ is model mismatch, and $\mathbf{s} = [h, \Theta, v, \omega]^\top \in \mathbb{R}^{10}$ with $h$ being $\text{CoM height}$, $v = [\text{CoM x-velocity, CoM y-velocity, CoM z-velocity}]^{\top}$, $\Theta = [\text{roll, pitch, yaw}]^\top$, and $w$ being the angular velocity of $\Theta$ in world coordinates. The $({\mathbf{A}}(\mathbf{s}(k)), {\mathbf{B}}(\mathbf{s}(k)))$ in \cref{realsyserror} is the available knowledge of physics model for designing HA-Teacher's action policy: 
\begin{align}
\mathbf{a}_{\text{HA}}(t) = \mathbf{F}_{k} \cdot \widehat{\mathbf{e}}(t) = \mathbf{F}_{k} \cdot (\mathbf{e}(t) - \mathbf{e}_{k}^*),  ~~~t \in \mathbb{T}^{\tau}_{k} \label{hapolocy}  
\end{align}

Hereto, we summarize HA-Teacher's working mechanism by considering \cref{aset4,center,realsyserror,hapolocy}. When the real-time states of the robot, controlled by HP-Student, move from safe self-learning space $\mathbb{L}$ to marginally-safe space $\mathbb{S}  \setminus  \mathbb{L}$, HA-Teacher uses the most recent sensor data, $\mathbf{s}(k)$, to update the physics model, denoted as $({\mathbf{A}}(\mathbf{s}(k)), {\mathbf{B}(\mathbf{s}(k))})$. This update facilitates the computation of both the real-time patch and the coupled action policy. The patch center, represented as $\mathbf{e}_{k}^* = \chi \cdot \mathbf{e}(k)$, is situated within HP-Student's self-learning space and aligns with HA-Teacher's real-time control objectives. This setup defines HA-Teacher's teaching task: to guide HP-Student towards an action policy that achieves high performance while ensuring safety at the safety boundaries.

Next, we present the design of HA-Teacher's action policy. Before proceeding, we outline a practical and common assumption regarding the model mismatch $\mathbf{h}(\mathbf{e}(k))$.  
\begin{assumption}
The model mismatch in $\mathbf{h}(\cdot)$ in \cref{realsyserror} is locally Lipschitz in ${\mathbb{P}_{t}}$ \eqref{aset4}, i.e., 
\begin{align}
&{( {{\bf{h}^\top}({{\bf{e}}_1}) - {\bf{h}}({{{\bf{e}}_2}})})^\top} \! \cdot {\mathbf{P}}_{{k}} \cdot ({\bf{h}}({{{\bf{e}}_1}) - {\bf{h}}( {{{\bf{e}}_2}})}) \nonumber\\
&\le \kappa  \cdot {( {{{\bf{e}}_1} - {{\bf{e}}_2}})^\top} \! \cdot {\mathbf{P}}_{{k}} \cdot ( {{{\bf{e}}_1} - {{\bf{e}}_2}}),~{\mathbf{P}}_{{k}} \succ 0,~\forall {\mathbf{e}_1}, {\mathbf{e}_2} \in {\mathbb{P}_{t}}. \nonumber
\end{align}
\label{assm11}
\end{assumption}
\vspace{-0.5cm}
We present the design of HA-Teacher's safety-assured action policy in the following theorem. Due to page limit, its detailed proof is provided in \cite{proof}.
\begin{theorem}\cite{proof}
Consider the self-learning space \eqref{aset3} and the real-time patch \eqref{aset4} with its patch center \eqref{center}, with their parameters satisfying: 
\begin{align}  
\theta + \chi \cdot \eta < 1. \label{pcon1}
\end{align}
Meanwhile, compute the matrix ${\mathbf{F}}_{k}$ in HA-Teacher's action policy in \cref{hapolocy} according to 
\begin{align}
{\mathbf{F}}_{k} = {\mathbf{R}}_{k} \cdot {\mathbf{Q}}_{k}^{-1}, \hspace{1cm} {\mathbf{P}}_{k} = {\mathbf{Q}}_{k}^{-1}, \label{acc0}
\end{align}
with ${\mathbf{R}}_{k}$ and ${\mathbf{Q}}_{k}$ satisfying: 
\begin{align}
&\left[ {\begin{array}{*{20}{c}}
{\mathbf{Q}}_{k} &  {\mathbf{R}}_{k}^\top\\
{\mathbf{R}}_{k} & \mathbf{T}_{k}
\end{array}}\right] \succ 0,  \label{thop1} \\
&\mathbf{I} - \overline{\mathbf{C}} \cdot \mathbf{Q}_{k} \cdot \overline{\mathbf{C}}^{\top} \succ 0,\label{thop2} \\
&\mathbf{I} - \overline{\mathbf{D}} \cdot \mathbf{T}_{k} \cdot \overline{\mathbf{D}}^{\top} \succ 0,\label{thop3}\\
&\mathbf{Q}_{k} - n \cdot \textbf{diag}^{2}(\mathbf{s}(k)) \succ 0,\label{thop2pp} \\
&\left[\!\!\!{\begin{array}{*{20}{c}}
(\alpha \!-\!  (1 \!+\! \frac{1}{\gamma}) \cdot \kappa) \!\cdot\! \mathbf{Q}_{k} & \mathbf{Q}_{k} \cdot \mathbf{A}^\top(\mathbf{s}(k)) \!+\! \mathbf{R}_{k}^\top \cdot \mathbf{B}^\top\\
\mathbf{A}(\mathbf{s}(k)) \cdot \mathbf{Q}_{k} \!+\! \mathbf{B} \cdot \mathbf{R}_{k} & (1 \!+\! \gamma)^{-1} \!\cdot\! \mathbf{Q}_{k} 
\end{array}}\!\!\!\right] \succ 0,\label{thop4}
\end{align}
where $\overline{\mathbf{C}} \triangleq \mathbf{C} \cdot \textbf{diag}^{-1}\{\theta \cdot \mathbf{c}\}$,  $\overline{\mathbf{D}} = \mathbf{D} \cdot \textbf{diag}^{-1}\{\mathbf{d}\}$, $\gamma > 0$, $0 < \alpha < 1$, and $\kappa$ is the Lipschitz bound given in \cref{assm11}. According to \cref{csafe}, HA-Teacher's action policy has assured safety. 
\label{thm10007p}
\end{theorem}

We note that using the CVX toolbox \cite{grant2009cvx}, the $\mathbf{Q}_{k}$ and $\mathbf{R}_{k}$ can be computed from \cref{thop1,thop2,thop3,thop2pp,thop4} for the $\mathbf{F}_{k}$.

\subsubsection{\textbf{HP-Student: Self-Learning and Teaching-to-Learn}}
HP-Student will learn from the HA-Teacher for a safe action policy in the teaching-to-learn space and engage in self-learning for a high-performance action policy in the self-learning space. The integration of these two learning paradigms enables the HP-Student to achieve a safe and high-performance action policy. The teaching horizon of the HA-Teacher is a crucial design. Specifically, during the teaching period, HA-Teacher generates a sequence of safe experience tuples, stored in HP-Student’s replay buffer for safety learning, as illustrated in \cref{learningframework}. This allows HP-Student to progressively internalize safety constraints, ensuring its learned policies adhere to safe operational bounds.

If the HA-Teacher's action policy cannot guide the robot to return to the self-learning space for a high-performance action policy (as illustrated by $\mathbb{P}_{k4}$ in \cref{phasewithpatch}), 'Teach-to-Learn' will take precedence over `Self-Learn' solely for the sake of learning safety. We note Theorem \ref{thm10007p} presents a design for a safety-assured action policy; however, the computing of the teaching horizon remains open. Hence, we offer guidance on the reasonable teaching horizons, formally stated in the following theorem. Due to page limit, its proof is in \cite{proof}.

\begin{theorem}[\textbf{Teaching Horizon} \cite{proof}]
Given parameters $\epsilon > 0$, $0< \chi < 1$, and $0< \alpha < 1$, if HA-Teacher's teaching horizon $\tau$ satisfies: 
\begin{align}
\tau \ge \tau_{\min} \triangleq  {\frac{{\ln(1 \!-\! (1\!+\!\epsilon) \cdot \chi^{2})  \!-\! 2\ln(1+\epsilon^{-1}) \!-\! \ln(1\!+\!\chi) }}{{\ln \alpha}}}, \nonumber
\end{align} 
we have ${\bf{e}}(k + \tau) \in \mathbb{L}$, i.e., the robot states can come back to HP-Student's self-learning space.   
\label{guidance}
\end{theorem}

With the teaching horizon in sight, we can proceed to present the self-learning and teaching-to-learn.

\noindent \underline{\textbf{Self-Learning}}. As shown in \cref{learningframework}, HP-Student adopts an \textit{actor-critic} architecture \cite{lillicrap2015continuous, haarnoja2018softactorcriticoffpolicymaximum} in DRL, with an experience replay buffer to enhance sample efficiency and mitigate temporal correlations. Its observation space is structured as: $\mathbf{o}_t = [\mathbf{z}_t, \mathbf{e}_t]^T$, where $\mathbf{z}_t$ is the exteroceptive observation, and $\mathbf{e}_t$ is the proprioceptive tracking error: $\mathbf{e}_t \triangleq \mathbf{s}_t - \mathbf{s}_{d}$, with $\mathbf{s}_d$ being the desired state. The self-learning objective is to optimize the action policy $\mathbf{a}_{\text{HP}}(k) = \pi(\mathbf{o}(k))$ which maximizes the expected return from the initial state distribution:
\begin{align}
&{\mathcal{Q}^\pi}( {\mathbf{e}(k), \mathbf{a}_{\text{HP}}(k)}) \nonumber\\
&\hspace{1.6cm} = {\mathbf{E}_{\mathbf{e}(k) \sim \mathbb{L}}}\left[ {\sum\limits_{t = k}^\infty {{\gamma^{t-k}} \cdot \mathcal{R}\left(\mathbf{e}(t), \mathbf{a}_{\text{HP}}(t) \right)}} \right]\!, \label{bellman}
\end{align}
where $\mathbb{L}$ denotes the self-learning space defined in \cref{aset3}, $\mathcal{R}(\cdot)$ is the reward function that maps the state-action pairs to real-value rewards, $\gamma \in [0,1]$ is the discount factor, balancing the immediate and future rewards. The expected return (\ref{bellman}) and action policy $\pi(\cdot)$ are parameterized by the critic and actor networks, respectively. The reward function of HP-Student is designed to enhance task-oriented performance in robots through runtime learning. Examples include minimizing travel time and power consumption in navigation and ensuring accurate state tracking in locomotion.

\setlength{\customwidth}{0.2078\textwidth}
\setlength{\fighspace}{-.241cm}

\begin{figure*}[http]
    \begin{subfigure}[b]{0.37\textwidth}
        \centering
        \includegraphics[width=\linewidth]{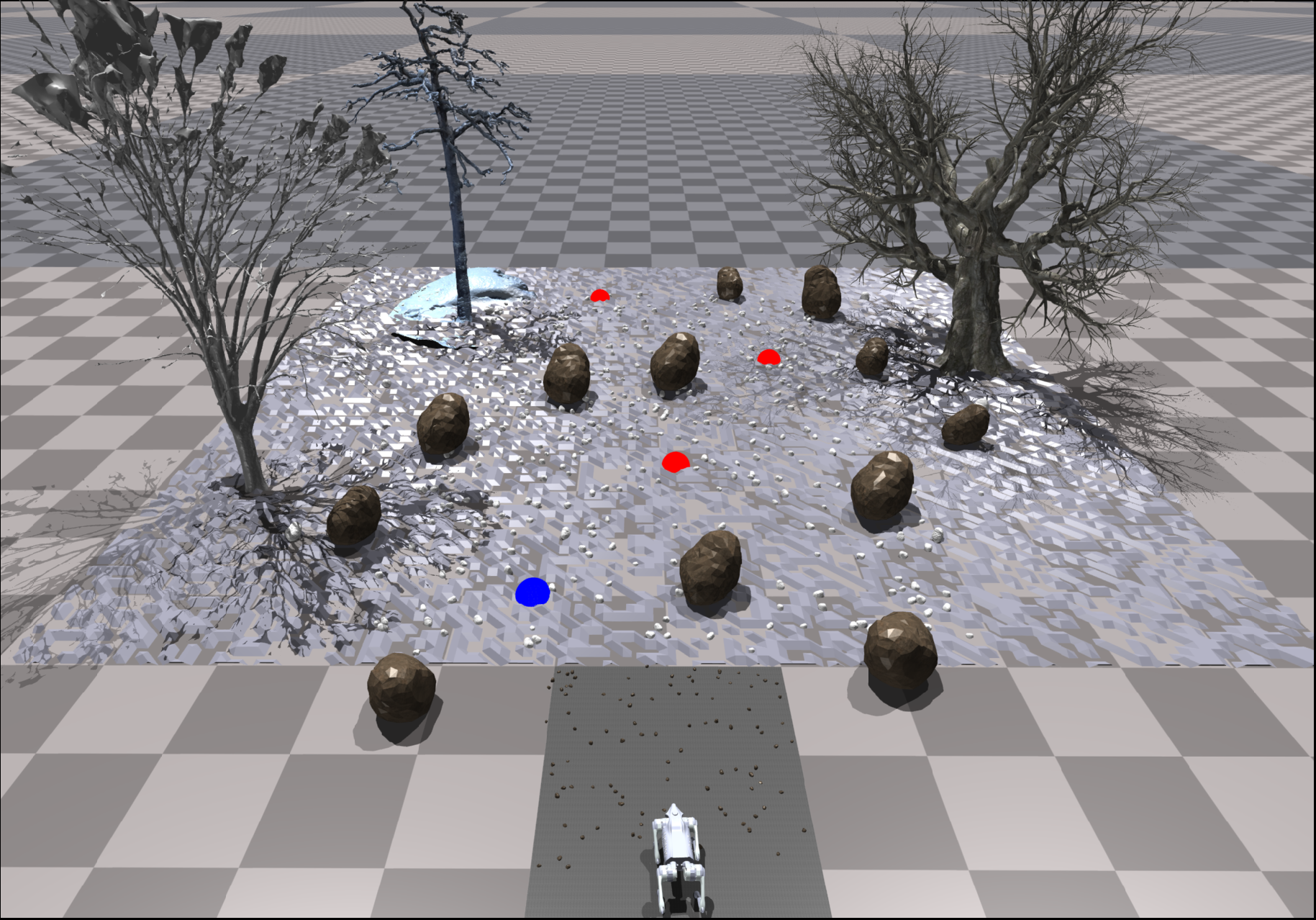}
        \caption{Dynamic wild environments in Isaac Gym}
    \end{subfigure}
    \label{sceneoverview}
    \hspace{\fighspace}
    \begin{subfigure}[b]{\customwidth}
        \centering
        \includegraphics[width=\linewidth]{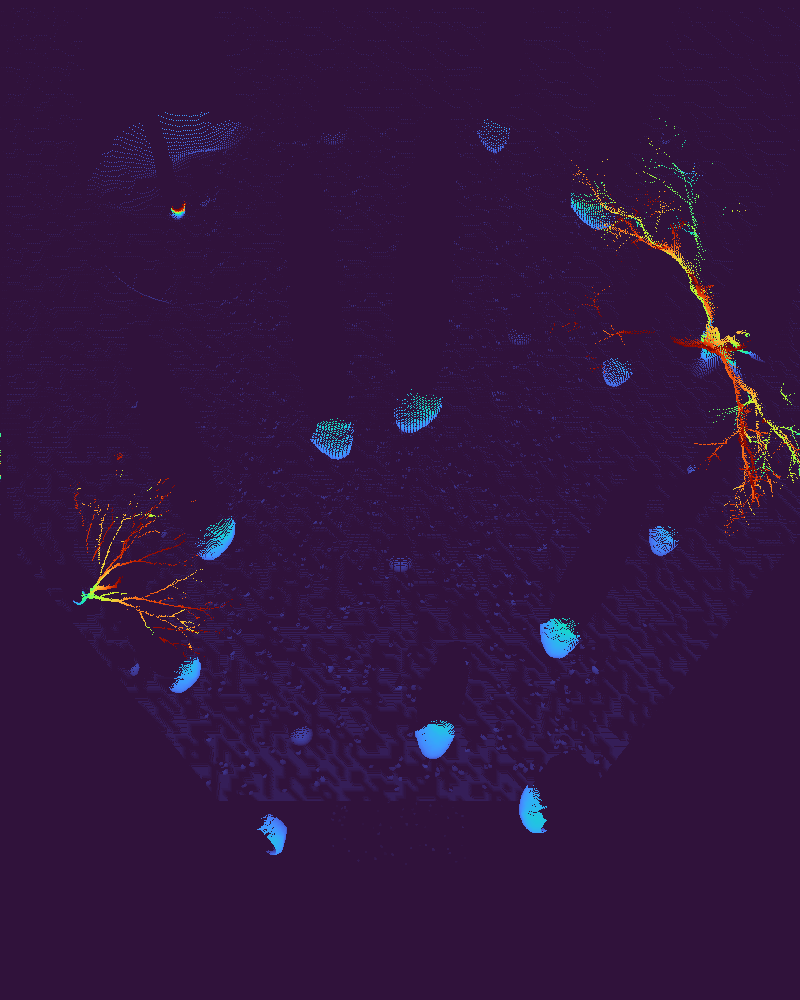}
        \caption{Raw BEV Map}
    \end{subfigure}
    \label{bevmap}
    \hspace{\fighspace}
    \begin{subfigure}[b]{\customwidth}
        \centering
        \includegraphics[width=\linewidth]{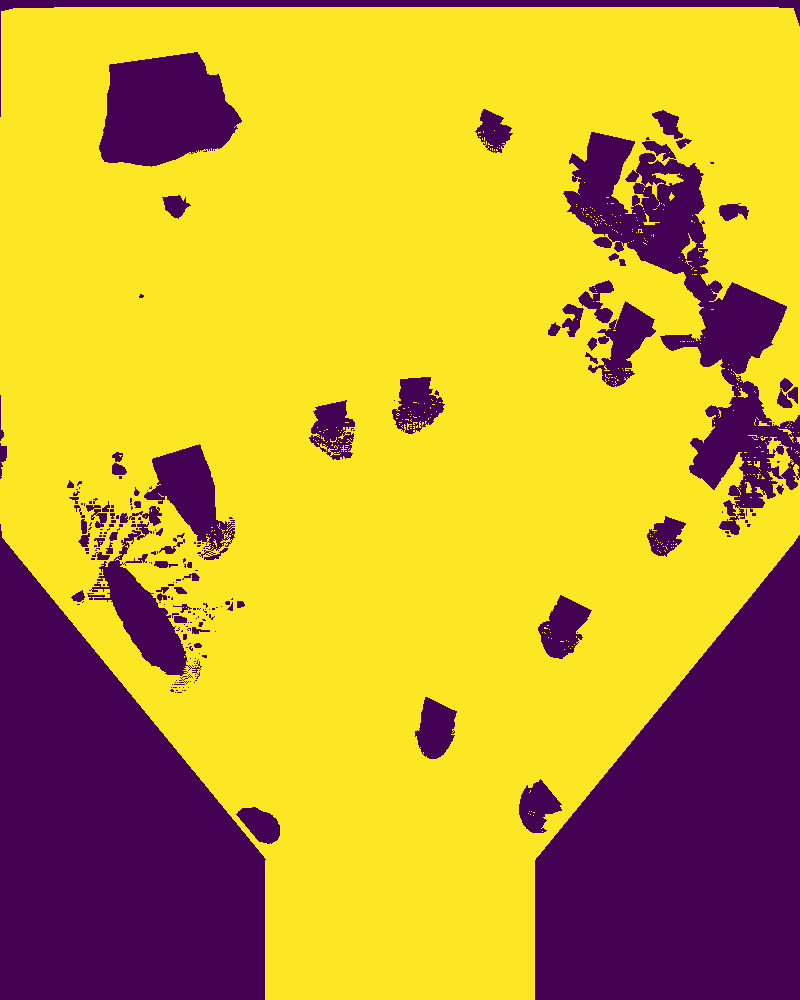}
        \caption{Occupancy Map}
    \end{subfigure}
    \label{occupancymap}
    \hspace{\fighspace}
    \begin{subfigure}[b]{\customwidth}
        \centering
        \includegraphics[width=\linewidth]{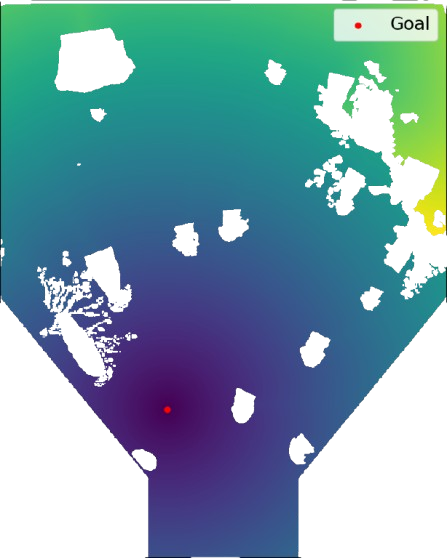}
        \caption{Cost Map}
    \end{subfigure}
    \label{costmap}
    \vspace{-0.0cm}
\vspace{-.45cm}
\caption{ Fig (a) is an overview of the dynamic wild environments: transiting from the flat terrain to unstructured and uneven ground. The quadruped robot navigates to the destination following the waypoints with minimum costs through runtime learning. The next waypoint is highlighted in \textcolor{blue}{blue}, and the remaining waypoints are marked in \textcolor{red}{red}. Figs. (b), (c) and (d) illustrate the cost map generation pipeline using the point cloud data.}
\label{ep5}
\end{figure*}

\noindent \underline{\textbf{Teaching-to-Learn}}. HP-Student controls the robot in normal situation. If its action $\mathbf{a}_{\text{HP}}(t)$ is unsafe by \cref{hpsafe}, HA-Teacher intervenes to ensure safe locomotion and generates a series of experience tuples over the teaching horizon $\tau$: 
\begin{align}
\mathrm{E}_{\tau} = \left\{\mathbf{o}(t), ~\mathbf{a}_{\text{HA}}(t), ~\mathbf{o}({t+1}), \mathcal{R}\left(\mathbf{o}(t), \mathbf{a}_{\text{HA}}(t) \right) \right\}^{k+\tau}_{t=k} \label{safedata}
\end{align}
which are continuously stored in HP-Student's replay buffer and uniformly sampled for safety learning. 

\begin{remark}
We note the action policy designed by the HA-Teacher in \cref{thm10007p}, is inherently safety-assured. The group of tuples $\mathrm{E}_{\tau}$ in \cref{safedata} documents the HA-Teacher's successful experiences in controlling a robot from the safety boundary to a secured self-learning space. Consequently, the HP-Student will learn from these HA-Teacher's experience,  specifically on how to safely manage the robot at the safety boundary. This approach enables the HP-Student to develop a safe and high-performance action policy suitable for real-time operations in dynamics wild environments. 
\end{remark}

\section{EVALUATION}
We utilize Nvidia Isaac Gym \cite{isaacgym} and the Unitree Go2 robot to evaluate our proposed runtime learning framework. In Isaac Gym, we create dynamic wild environments, which include various natural elements, such as unstructured terrain, movable stones, and obstacles like trees and large rocks. Additionally, we arrange multiple waypoints at reasonable intervals across the terrain to simulate real-world tasks for the robot, such as outdoor exploration and search-and-rescue operations. The aim is to guide the robot to its destination by sequentially following these waypoints  while minimizing traversal costs and adhering to safety constraints. The established wild environments can be found in \cref{ep5} (a).

\subsection{HP-Student: Task-Oriented Reward}
The reward for the DRL agent (i.e., our HP-Student) primarily focuses on three key aspects: robot safety and stability, travel time, and energy efficiency.

\subsubsection{Stability}
Stability is essential for the robot's safe locomotion control. Thus we consider a Lyapunov-like reward in \cite{westenbroek2022lyapunov, caophysics}, which incorporates both safety and stability:
\begin{align}
r_{stability} = \mathbf{e}^{\top}(t) \cdot \mathbf{P} \cdot \mathbf{e}(t) - \mathbf{e}^{\top}(t\!+\!1) \cdot \mathbf{P} \cdot \mathbf{e}(t\!+\!1), \label{stability}
\end{align}
The computation of $\mathbf{P}$ can follow the guidance in \cite{caophysics}. 

\subsubsection{Travel Cost}
The Euclidean distance between the robot and the waypoint is defined as:
\begin{align}
d(t) = \|\hat{\mathbf{x}}_{b}(t) -\hat{\mathbf{p}}_{wp}\|_2  \label{distance_wp}
\end{align}
where $\hat{\mathbf{p}}_{wp}$ is the location of next waypoint, and $\hat{\mathbf{x}}_{b}$ is the position of the robot base in the world frame.
Inspired by \cite{geng2020deepreinforcementlearningbased} and \cite{minimumtime}, 
the navigation reward is defined as:
\begin{align}
r_{nav}(t) = c_1 \cdot r_{dis}(t) +~c_2 \cdot r_{wp} + r_{obs} \label{travel}
\end{align}
where $r_{dis}(t) = d(t) - d(t-1)$ is the reward for forwarding the waypoint. $r_{wp} = e^{-\lambda \cdot T_{reach}}$ rewards the robot as it reaches the waypoint. $\lambda \in (0,1)$ is a time decay factor and $T_{reach}$ denotes the time step when the waypoint is reached. 
$r_{obs}$ serves as a penalty when the quadruped collides with the obstacles. $c_1$, $c_2$ and $c_3$ are used hyperparamters.


\begin{figure*}[htbp]
    \centering
        \begin{subfigure}[b]{0.325\textwidth}
        \centering
        \includegraphics[width=\linewidth, trim=10 30 70 90, clip]{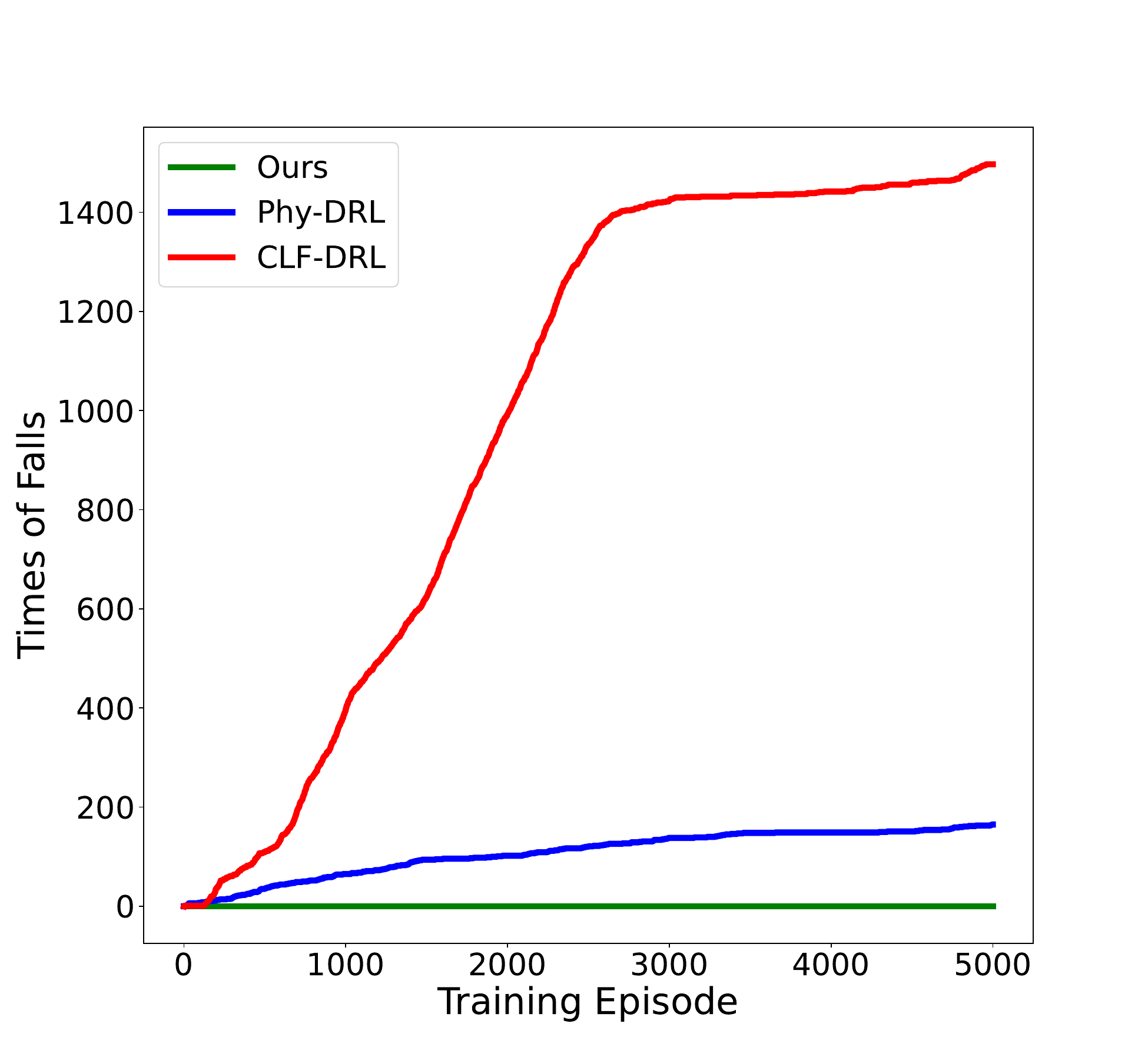}
        \caption{Accumulated number of robot's falls}
        \label{fig:reward}
    \end{subfigure}
    \hfill
    \begin{subfigure}[b]{0.325\textwidth}
        \centering
        \includegraphics[width=\linewidth, trim=10 30 70 90, clip]{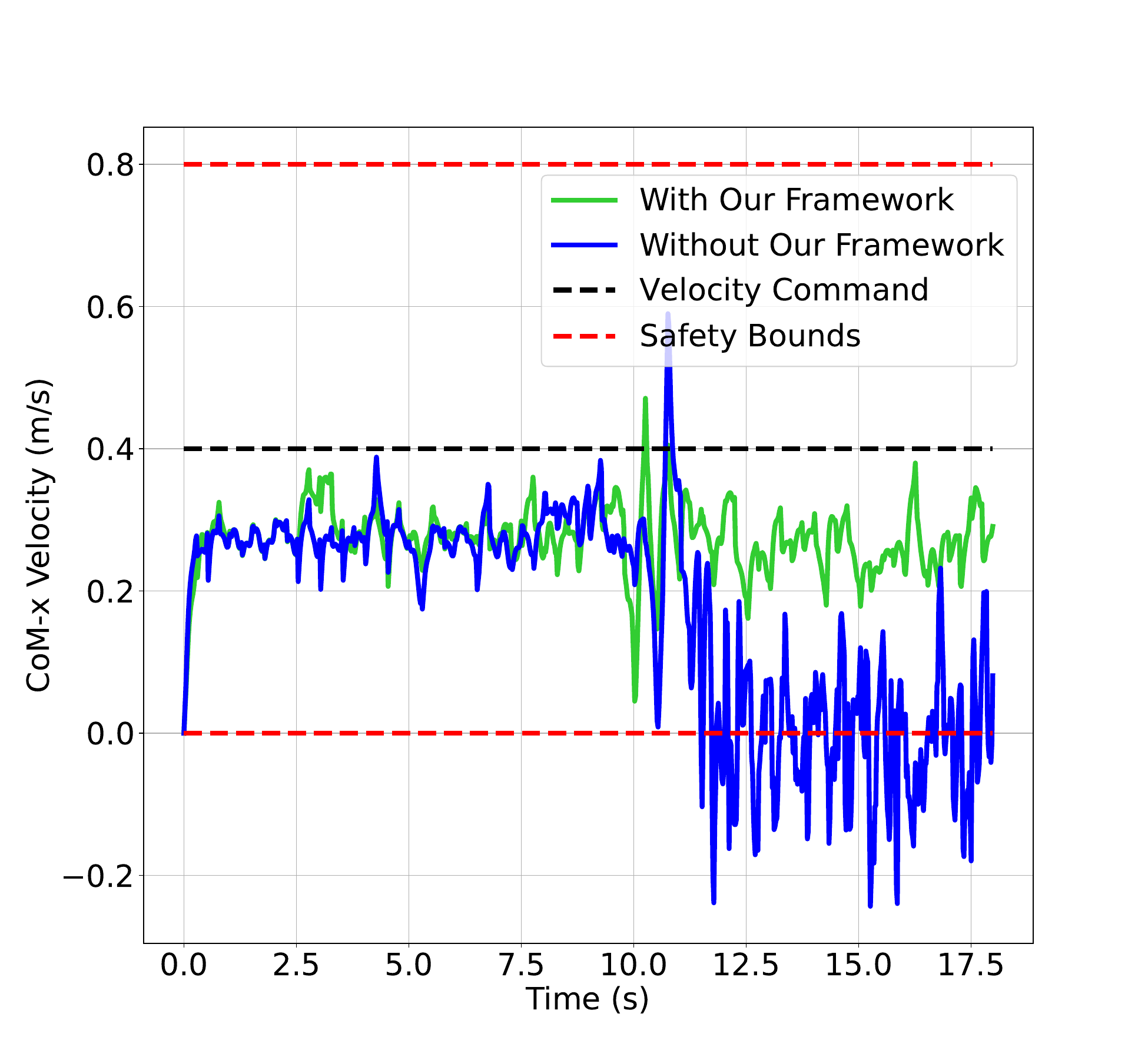}
        \caption{Trajectory of the robot's CoM velocity}
        \label{fig:velocity}
    \end{subfigure}
    \hfill
    \begin{subfigure}[b]{0.325\textwidth}
        \centering
        \includegraphics[width=\linewidth, trim=10 30 70 90, clip]{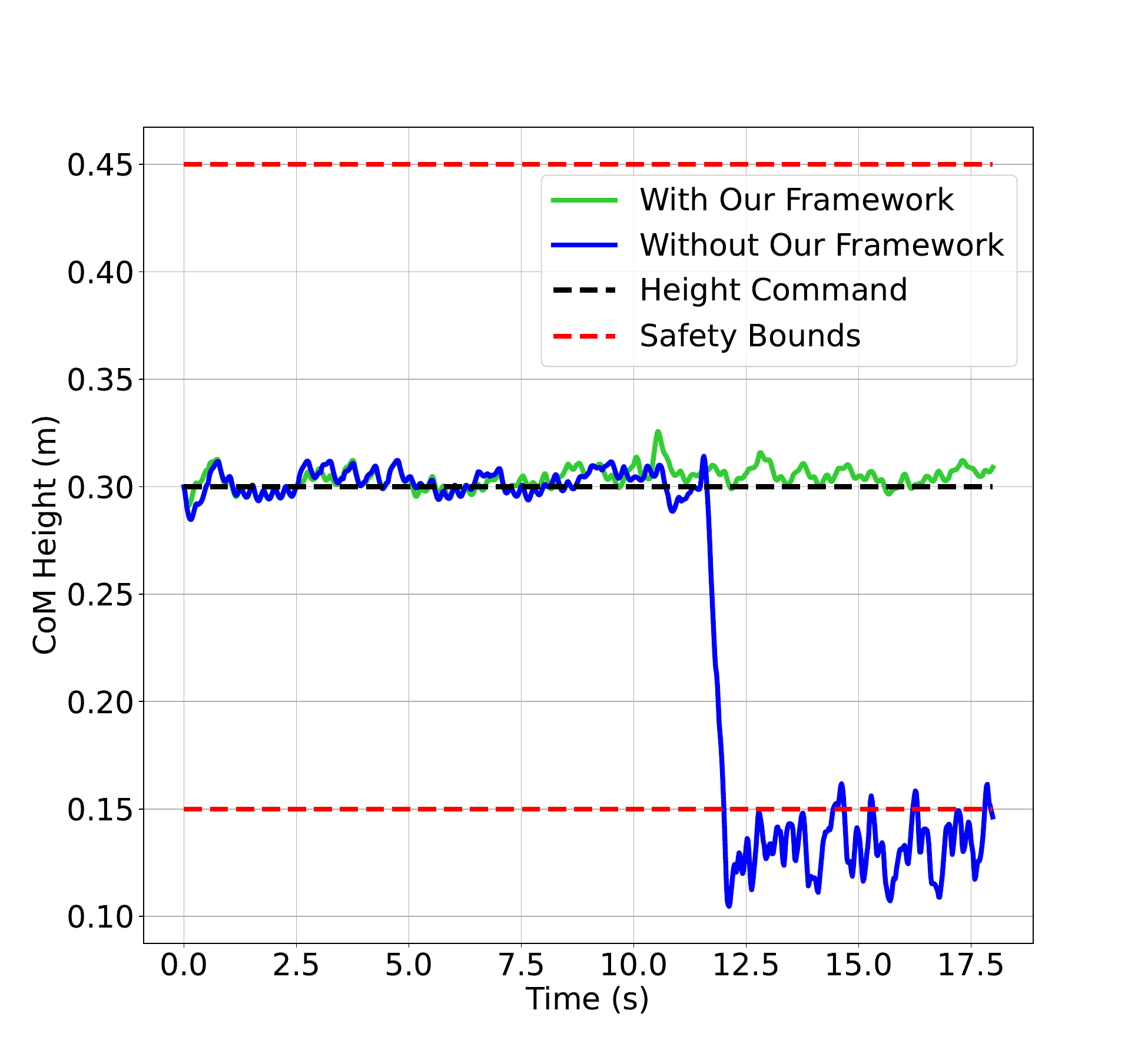}
        \caption{Trajectory of the robot's CoM height}
        \label{fig:height}
    \end{subfigure}
    \caption{(a) is the accumulated number of robot falls over learning episodes. (b) and (c) depict the robot's CoM trajectories, highlighting the effectiveness of our framework in ensuring runtime learning safety under dynamic wild environments.}
    \label{fig:rtl}
\end{figure*}

\subsubsection{Energy Consumption}
Power efficiency remains a major challenge for robots in outdoor settings. We model the motor as a non-regenerative braking system \cite{yang2022fast}:
\begin{align}
p_{motor} = \max \{\underbrace{\tau_{m} \cdot \omega_{m}}_{output\ power} + \underbrace{L_{copper} \cdot \tau_{m}^2}_{heat\  dissipation},~0\} \label{energy}
\end{align}
where $\tau_m$ and $\omega_m$ are motor's torque and angular velocity respectively. $L_{copper}$ is copper loss coefficient.
Taking $c_m$ as a hyperparameter, we define the energy consumption reward:
\begin{align}
r_{energy} = -c_m \cdot p_{motor} \label{energyrewardadd}
\end{align}

The ultimate reward function, designed to guide the HP-Student in learning a safe and high-performance policy as defined in \cref{bellman}, is given by integrating the components from above \cref{stability,travel,energyrewardadd}, i.e., $\mathcal{R}\left(\mathbf{e}(t), \mathbf{a}_{\text{drl}}(t) \right) = r_{stability} +  r_{nav} + r_{energy} + \widehat{c}  \cdot r_{aux}$, where $r_{aux}$ denotes the auxiliary reward for tracking velocity and orientation, with a small coefficient $\widehat{c}$ to promote smooth locomotion.

HP-Student's exteroceptive observation includes the Euclidean distance $d_{wp}$ between robot and next waypoint, and the robot's heading angle deviation to the waypoint ${\psi}_{wp}$. Its priproceptive tracking error is shared with HA-Teacher.

\subsection{HA-Teacher: Safety Critical Design}
The real-time physics-model knowledge in \cref{realsyserror} -- used for HA-Teacher design in \cref{thm10007p} -- is obtained by considering the robot's dynamics as described in \cite{di2018dynamic}: 
\begin{align} 
\mathbf{B}(\mathbf{s}(k))  &= \left[ {\begin{array}{*{20}{c}}
{{\mathbf{O}_3}}&{{\mathbf{O}_3}}&{{\mathbf{O}_3}}&{{\mathbf{O}_3}}\\
{{\mathbf{O}_3}}&{{\mathbf{O}_3}}&{{\mathbf{O}_3}}&{{\mathbf{O}_3}}\\
{{\mathbf{O}_3}}&{{\mathbf{O}_3}}&{T \cdot {\mathbf{I}_3}}&{{\mathbf{O}_3}}\\
{{\mathbf{O}_3}}&{{\mathbf{O}_3}}&{{\mathbf{O}_3}}&{T \cdot {\mathbf{I}_3}}
\end{array}} \right]\!\!, \nonumber\\
\mathbf{A}(\mathbf{s}(k))  &= \left[\!\! {\begin{array}{*{20}{c}}
{{1}}&{{\mathbf{O}_{1 \times 5}}}&{T}&{{\mathbf{O}_{1 \times 3}}}\\
{{\mathbf{O}_3}}&{{\mathbf{I}_3}}&{{\mathbf{O}_3}}&T \cdot {\mathbf{R}(\phi(k),\theta(k),\psi(k))}\\
{{\mathbf{O}_3}}&{{\mathbf{O}_3}}&{{\mathbf{I}_3}}&{{\mathbf{O}_3}}\\
{{\mathbf{O}_3}}&{{\mathbf{O}_3}}&{{\mathbf{O}_3}}&{{\mathbf{I}_3}}
\end{array}} \!\!\right]\!\!,  \nonumber 
\end{align}
where ${\mathbf{R}(\phi(k),\theta(k),\psi(k))}$ is the real-time rotation matrix and $T$ is the sampling period of the robot system.




    
    

\subsection{Experimental Results}
\subsubsection{Safety Assessment}
We set quadruped's desired state to be: $v_x^d = $ 0.4 m/s, $h_d$ = 0.3 m,  $\omega_z^d = \omega_z^{ref}$, where $\omega_z^{ref}$ is the reference angular velocity trajectory generated by the planner. Following the definition in \cref{aset2}, we certify the robot's safety set: 
\begin{align}
    \mathbb{S} = \{\mathbf{e} \in \mathbb{R}^{10}~|~|e_{v_x}| \le 0.4~\text{m/s}, |e_h| \le 0.15~\text{m}\}, 
    \label{exp_safetyset}
\end{align} 
where $e_{v_x} \triangleq v_x - v_x^d$ denotes the error between the robot's forwarding velocity and its command, and $e_h = h - h_d$ is the error between the robot's CoM height and its desired height. The corresponding action space in \cref{org} is:  
\begin{align}
    \mathbb{A} = \{\mathbf{a} \in \mathbb{R}^{6}~|~|\mathrm{a}_{v}| \le 10~\text{m/$s^2$}, |\mathrm{a}_{\omega}| \le 20 ~\text{rad/$s^2$}\},
    \label{exp_safetyset}
\end{align} 
where $\mathrm{a} = [\mathrm{a}_v,~ \mathrm{a}_{\omega}]^\top$, with $\mathrm{a}_v \in \mathbb{R}^3$ and $\mathrm{a}_{\omega} \in \mathbb{R}^3$ denoting the robot's linear and angular acceleration, respectively.

To establish the self-learning space $\mathbb{L}$ in \cref{aset3}, we choose $\eta = 0.7$.
And $\omega_z^{ref}$ is clipped within a range to ensure smooth angular velocity during navigation: $\omega^{ref}_z \in [-0.7, 0.7]$ rad/s. For HA-Teacher, we select $\alpha = 0.8$, $\kappa = 0.01$, $\chi = 0.15$, and $\tau = 10$ satisfying condition in \cref{guidance}. By utilizing the CVX toolbox \cite{grant2009cvx}, we can compute $\mathbf{F}_{k}$ from \cref{thop1,thop2,thop3,thop2pp,thop4} in \cref{thm10007p} for HA-Teacher's action policy \eqref{hapolocy}.

The experiment also includes comparisons with the state-of-the-art safe DRL frameworks: CLF-DRL \cite{westenbroek2022lyapunov} and Phy-DRL \cite{caophysics}. We assess the safety assurance of our framework, CLF-DRL, Phy-DRL, and sole HA-Teacher, by analyzing the number of falls during runtime learning, as summarized in \cref{table:exp_result} and \cref{fig:rtl} (a). A demonstration video is available at: \href{https://www.youtube.com/shorts/2IsZQYwjccg}{\textcolor{blue}{https://youtube.com/shorts/2IsZQYwjccg}}


The demonstration video, in conjunction with \cref{table:exp_result} and training reward in \cref{episodicreturns}, highlights the effectiveness of our approach in ensuring runtime safety for the quadruped robot during navigation tasks in the dynamic wild environments. Additionally, the corresponding trajectories of the robot, depicted in \cref{fig:rtl}, further illustrate the capability of our framework in maintaining the robot states within the safety set, as outlined in \cref{exp_safetyset}, throughout runtime learning.

\begin{table*}[h]
    \centering
    \renewcommand{\arraystretch}{1.05}  
    \resizebox{1.0\textwidth}{!}{ 
    \begin{tabular}{c|c|c|cccc|cc}
        \toprule
        \multirow{3}{*}{\textbf{Methodologies}} & \multirow{3}{*}{\textbf{Model -- ID}} &
        \multicolumn{5}{c|}{\textbf{Navigation Performance}} &
        \multicolumn{2}{c}{\textbf{Energy Efficiency}} \\
        \cmidrule(lr){3-7} \cmidrule(lr){8-9} & & \textbf{Success} & \textcolor{red}{\textbf{Is Fall}} & \textcolor{red}{\textbf{Collision}} & \textcolor{blue}{\textbf{Num (wp)}} & \textbf{Travel Time (s)} & \textbf{Avg Power (W)} & \textbf{Total Energy (J)} \\
        \midrule
        \multirow{3}{*}{\textbf{CLF-DRL}}  
        & clfdrl-ep-1500  & \text{No}  & \textcolor{red}{\text{Yes}} &  \textcolor{red}{\text{Yes}} & \textcolor{blue}{\text{0}}  & \text{N/A} & \text{N/A} & \text{N/A} \\ 
        & clfdrl-ep-3000 & \text{No}  & \textcolor{red}{\text{Yes}} &  \textcolor{red}{\textbf{No}} & \textcolor{blue}{\text{0}} & \text{N/A} & \text{N/A} & \text{N/A} \\
        & clfdrl-ep-4500  & \text{No}  & \textcolor{red}{\text{Yes}} &  \textcolor{red}{\textbf{No}} & \textcolor{blue}{\text{1}} & \text{N/A} & \text{N/A} & \text{N/A}\\
        \midrule
        \multirow{3}{*}{\textbf{Phy-DRL}}  
        & phydrl-ep-1500  & \text{No}  & \textcolor{red}{\text{Yes}} & \textcolor{red}{\textbf{No}}  & \textcolor{blue}{\text{0}}  & \text{N/A} & \text{N/A} & \text{N/A} \\
        & phydrl-ep-3000 & \text{No}  & \textcolor{red}{\textbf{No}} & \textcolor{red}{\text{Yes}}  & \textcolor{blue}{\text{2}}  & $\infty$ & \text{504.3827} & $\infty$ \\
        & phydrl-ep-4500 & \text{No}  & \textcolor{red}{\textbf{No}} &   \textcolor{red}{\textbf{No}} & \textcolor{blue}{\textbf{4}} & 56.6316 & 489.5142 & 27721.97 \\
        \midrule
        \multirow{3}{*}{\textbf{Ours}}  
        & rtl-ep-1500  & \textbf{No}  & \textcolor{red}{\textbf{No}} & \textcolor{red}{\text{Yes}} & \textcolor{blue}{\text{2}}  & $\infty$ & \text{491.7283} & $\infty$ \\
        & rtl-ep-3000  & \textbf{Yes}  & \textcolor{red}{\textbf{No}} & \textcolor{red}{\textbf{No}} & \textcolor{blue}{\textbf{4}}  & 48.6417 & \textbf{488.5232} & \text{23762.59} \\
        & rtl-ep-4500  & \textbf{Yes}  & \textcolor{red}{\textbf{No}} & \textcolor{red}{\textbf{No}}  &  \textcolor{blue}{\textbf{4}} & \textbf{45.3792} & \text{490.9204} & \textbf{22277.53} \\
        \midrule
        \textbf{HA-Teacher} & -- & \textbf{Yes}  & \textcolor{red}{\textbf{No}} &  \textcolor{red}{\textbf{No}} & \textcolor{blue}{\textbf{4}} & 59.2706  & \text{493.8499} & \text{29270.78}\\
        \bottomrule
    \end{tabular}
    }
    \caption{Comparison of Different DRL Agents, the Sole HA-Teacher, and Our Runtime Learning Framework. \textcolor{red}{Safety-related} metrics are highlighted in red, while key navigation metrics are marked in \textcolor{blue}{blue}. In instances where the robot falls, other task-related metrics are labeled as N/A due to the safety-critical nature of the robot. For collision scenarios, travel time and energy consumption are represented as $\infty$, with power reflecting the average energy consumption prior to the collision.}
    \label{table:exp_result}
\end{table*}

\subsubsection{Performance Assessment}
\begin{figure}[htbp]
\begin{center}
\includegraphics[width=0.417\textwidth, trim=0 30 30 100, clip]{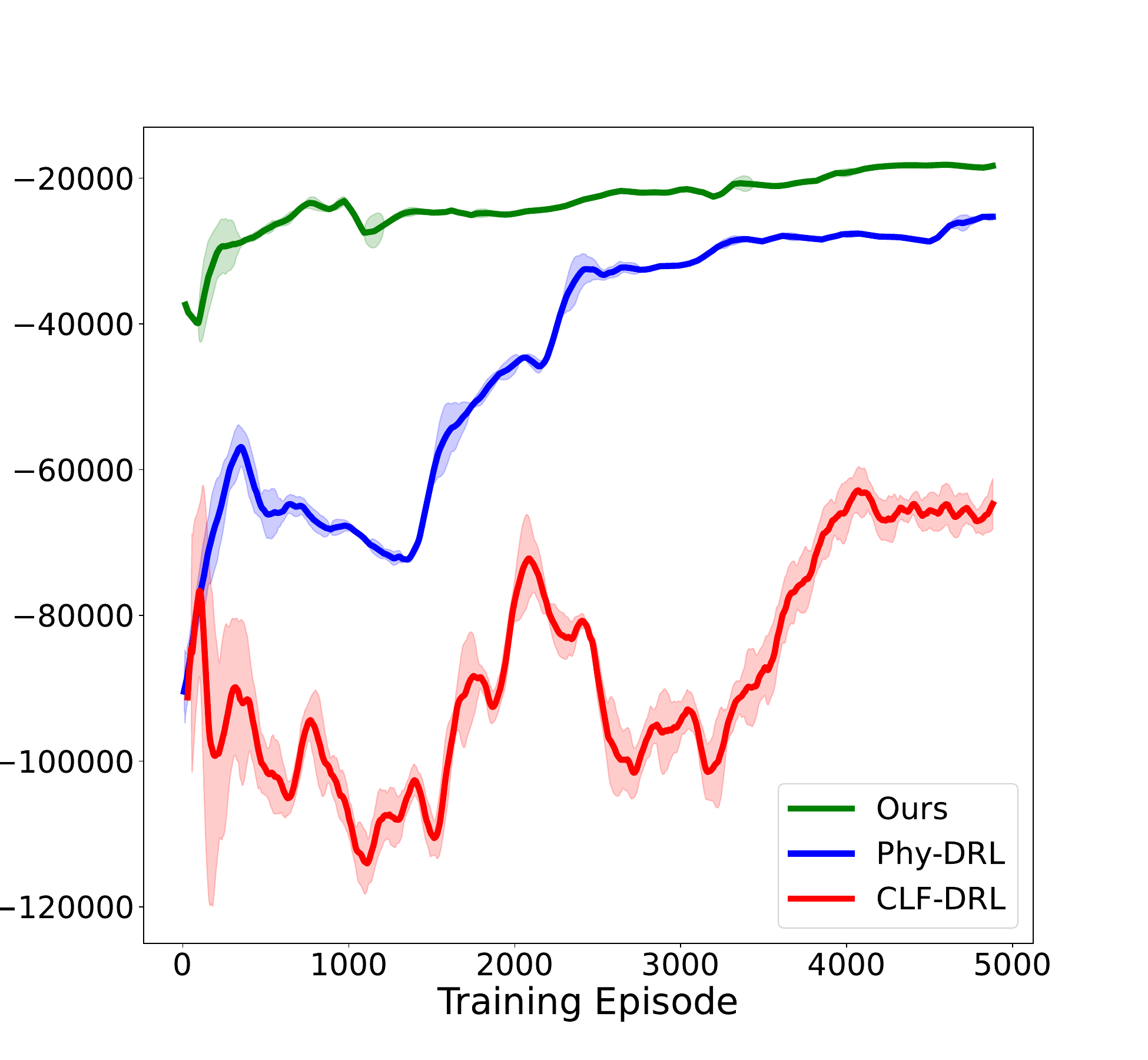}
\end{center}
\vspace{-0.2cm}
\caption{Episodic Returns During Runtime Learning}
  \label{episodicreturns}
  \vspace{-0.2cm}
\end{figure}

We assess the efficiency of our framework by comparing it against different DRL agents, i.e., Phy-DRL and CLF-DRL, and sole HA-Teacher, as summarized in \cref{table:exp_result}. The episodic returns are shown in \cref{episodicreturns} and models are selected after training 1500, 3000, and 4500 episodes for running the task. For navigation performance, we measure \textit{Success}--whether the robot reaches the destination, \textit{Is Fall}--whether the robot falls, \textit{Collision}--whether the robot collides with obstacles, \textit{Num (wp)}--number of waypoints the robot followed and \textit{Travel Time}. For energy efficiency, we evaluate \textit{Avg Power}--average motor power and \textit{Energy Consumption}--total energy used.

In \cref{table:exp_result}, CLF-DRL agent struggles to develop an optimal policy for navigation task. Benefiting from its inherent structure, Phy-DRL demonstrates improved performance but still faces slow convergence and safety issues during runtime learning. HA-Teacher ensures safe navigation, but it is inefficient in travel time and energy consumption. In contrast, under the same conditions, our framework enables the HP-Student to effectively adapt to the wild environments, achieving a safe and high-performance policy after runtime learning. The demonstration video is available at: \href{https://youtube.com/shorts/epw7sSYIiqs}{\textcolor{blue}{https://youtube.com/shorts/epw7sSYIiqs}}.


\setlength{\textfloatsep}{12pt}

\section{Conclusions and Future Work}
This paper presents a runtime learning framework that enables a quadruped robot to learn adaptively and safely in complex wild environments. The framework offers verifiable safety guarantees while enhancing learning performance during runtime adaptation. Experiments conducted with a Unitree Go2 robot in Isaac Gym have demonstrated the effectiveness of this runtime learning approach.

Moving forward, we will implement the proposed runtime learning framework on the real Unitree Go2 robot, 
enabling robot to safely learn in real-time physical environments.

\section{Acknowledgments}
This work was partly supported by the National Science Foundation under Grant CPS-2311084 and Grant CPS-2311085 and the Alexander von Humboldt Professorship Endowed by the German Federal Ministry of Education and Research.

\bibliographystyle{unsrt}
\bibliography{reference}
\end{document}